\documentclass{article}

 \usepackage[preprint]{neurips_2025}

\usepackage[utf8]{inputenc} 
\usepackage[T1]{fontenc}    
\usepackage{hyperref}       
\usepackage{url}            
\usepackage{booktabs}       
\usepackage{amsfonts}       
\usepackage{nicefrac}       
\usepackage{microtype}      
\usepackage{xcolor}         
\usepackage{natbib}
\usepackage{graphicx}
\usepackage{amsmath}
\usepackage{amssymb}
\usepackage{wrapfig}

\hypersetup{
    colorlinks=true,   
    linkcolor=[RGB]{15, 7, 166},   
    citecolor=[RGB]{15, 7, 166},   
    urlcolor=[RGB]{15, 7, 166}     
}

\title{Learning to Shard: RL for Co-optimizing the Parallelism Degrees and Per-operator Sharding Dimensions in Distributed LLM Inference}

\author{%
  Ruokai Yin\thanks{Correspondence to ruokai.yin@yale.edu. This work was completed during an internship at Microsoft.}\hspace{0.5mm} $^1$, Sattwik Deb Mishra$^2$,  Xuan Zuo$^2$, Hokchhay Tann$^2$,
  Preyas Shah$^2$,  Apala Guha$^2$\\
  $^1$Yale University, $^2$Microsoft Azure
}

\begin{document}

\maketitle

\begin{abstract}
Distributed LLM inference requires careful coordination of parallelization strategies across hundreds to thousands of NPUs to meet production SLOs. Current systems like Megatron-LM rely on static heuristics that separately configure parallelism degrees and per-operator sharding dimensions, leaving significant performance on the table as models scale and hardware topologies diversify. We introduce \textbf{Learn to Shard}, to our knowledge, the first RL-based approach to co-optimize both coarse-grained parallelism degrees and fine-grained per-operator sharding dimensions for distributed LLM inference. Our method employs an attention-based policy over an elite history that learns from high-performing strategies to efficiently navigate the vast combinatorial search space. Evaluated on H100 clusters with MoE models up to 1.6T parameters, Learn to Shard achieves up to 3.5$\times$ throughput improvement over metaheuristic baselines and 1.06$\times$ over Megatron heuristics.
\end{abstract}

\section{Introduction}

Large language models (LLMs), including LLaMA~\citep{touvron2023llama} and mixture-of-experts (MoE) models~\citep{liu2024deepseek}, have achieved remarkable performance across a wide range of tasks. 
However, their massive scale—often exceeding hundreds of billions or even trillions of parameters—demands inference over a vast number of NPUs to deliver production-quality latency and throughput. As a result, \emph{distributed inference} has become the default, and optimizing parallelism is critical for meeting service-level objectives (SLOs) under strict hardware and cost constraints.

Modern distributed inference relies on multiple types of parallelism, such as tensor parallelism (TP), expert parallelism (EP), and pipeline parallelism (PP), each offering distinct efficiency trade-offs. These strategies are not mutually exclusive: in fact, combining them into \emph{mixed parallelism} often yields better throughput and efficiency than any single type alone. In practice, existing inference systems like Megatron-LM~\citep{shoeybi2019megatron} determine the parallelization degrees of different parallelization types using heuristics and hand-tuned rules. 
We refer to the selection of parallelism degrees as the \emph{coarse-grained parallelization strategy}.

Beyond coarse-grained parallelization strategies, we emphasize a \emph{fine-grained} and largely overlooked strategy design axis—the selection of the \emph{per-operator sharding dimension}.
The per-operator sharding dimension specifies which tensor dimensions individual operators are sharded along. Existing inference systems fix this choice through static heuristics.
For instance, in Megatron’s TP implementation of MLP operators, FFN1 and FFN2 are sharded along feedforward dimension (dim 1 and 0 respectively).
While this minimizes the number of inter-NPU communications, it assumes fixed all-reduce operations that may be suboptimal on different interconnects or under strict latency constraints (e.g., all-gather often performs better). Exploring the sharding dimensions—especially together with parallelism degrees—remains a largely unexplored problem. 

With both the fine and coarse-grained parallelization strategies considered, the co-optimization space grows significantly. Specifically, as LLM models scale to trillions of parameters~\citep{nvidia2024gtc_slides}, hundreds of NPUs are required just for storing weights and KV caches, while thousands may be needed to meet inference SLOs. Therefore, heuristic approaches become increasingly unscalable. While empirical optimization continues to discover new parallelization strategies beyond existing heuristics~\citep{bhatia2025helix, qin2025chimera}, the time required to discover effective strategies increases as the search space expands. This motivates the need for an automated and scalable approach for searching for new strategies.

To automate parallelization strategy optimization, prior work has explored automated search via integer programming~\citep{lin2025uniap} and metaheuristic algorithms~\citep{raju2025cosmic}, offering various trade-offs between scalability and optimality. Reinforcement learning (RL) has recently emerged as a powerful tool in automating compiler optimizations~\citep{zhou2020transferable}, but its application to parallelization strategy remains largely unexplored. Furthermore, many other considerations such as intra-device tiling, collective algorithms, and model compression impact SLO and cost during inference. We ask whether RL-based agentic search is a feasible approach to this problem\footnote{In this work, we start the exploration by applying the RL-based agentic search to a subset of the problem (parallelization strategy)}.

In this work, we propose \textbf{Learn to Shard}, an RL-based agent that co-optimizes both the coarse-grained (parallelism degrees) and the fine-grained (per-operator sharding dimensions) parallelization strategy for distributed LLM inference. Our key insight is that high-performing parallelization strategies exhibit patterns that can be learned from the agent's explorations. Consequently, an RL-based approach becomes beneficial through its sampling efficiency. We develop an attention-based elite-context policy that learns from previously discovered strategy configurations to guide exploration in the vast and combinatorial optimization space—dramatically reducing the time-to-discovery compared to non-learning baselines.

We evaluate \textbf{Learn to Shard} on real-world workloads on H100 GPU clusters with up to thousands of devices. Our method achieves up to 3.5$\times$ throughput improvement over metaheuristic baselines, and outperforms Megatron-LM’s heuristic strategies by up to 1.06$\times$, requiring only 4000 search rounds out of $10^9$ possible configurations.



\section{Background and Related Work}

\begin{wrapfigure}{l}{0.5\textwidth}
\centering
\vspace{-15pt} 
\includegraphics[width=\linewidth]{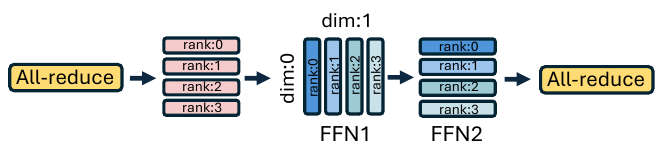}
\vspace{-20pt}
\caption{An illustrative example of the standard TP parallelism strategy.}
\vspace{-5pt}
\label{fig:background}
\end{wrapfigure}

\textbf{Parallelization types and strategies. } In tensor parallelism (TP), weight matrices for each layer are sharded and distributed across different NPUs, as shown in Figure~\ref{fig:background}. We only consider 1-D TP. Pipeline parallelism (PP) can be viewed as task parallelism at layer level, operator level, etc. In this work, we focus on layer level parallelism. In PP, different task subsets are placed on different NPUs, with adjacent pipeline stages transferring activations between devices. Expert parallelism (EP) is designed for modern MoE models, where different experts are placed on different NPUs and input tokens are routed to appropriate devices based on gating scores. Other parallel types also exist such as sequence parallelism and data parallelism which are complementary to our search framework and optimization space. 

For agentic search, we need a formal representation of the sharding strategy. We define a parallelization strategy with two parts: the coarse-grained and the fine-grained strategy. The coarse-grained strategy refers to the parallelization degrees, which refer to the number of NPUs that each specific parallelization type requires. In the TP-only example of Figure~\ref{fig:background}, the coarse-grained strategy can be described as \{TP=4, EP=1, PP=1\}.
The fine-grained strategy determines which dimension each operator's output tensor is sharded along. In the provided example, the fine-grained strategy for the loading operators of FFN1 and FFN2 is \{dim=1, dim=0\}.
In practice, the output tensor can be sharded on any dimension as long as correct communication collectives ensure dimension matching between producer and consumer operations. And we can represent arbitrary\footnote{Prior empirically discovered strategies are representable under this framework and thus discoverable through agentic search} sharding strategies in this way, for example, sequence parallelism~\citep{li2021sequence} shards the attention operators on the context length dimension\footnote{In this work, we allow sharding of attention operators on the head and hidden dimensions, and the sharding of MLP operators on hidden and feed-forward dimensions. We leave exploring context length sharding as a future work}.
Under this representation, the sharding strategy can be effectively modeled into a single set of integer values that can be seamlessly integrated into our agentic search flow.

\textbf{Related work. }
Extensive prior work has focused on automatically determining parallelism strategies for large model training and inference. 
Systems such as \textsc{Megatron-LM}~\citep{shoeybi2019megatron}, TensorFlow-XLA~\citep{openxla2025}, and \textsc{GSPMD}~\citep{xu2021gspmd} provide heuristic or rule-based parallelization strategies.
\textsc{Alpa}~\citep{zheng2022alpa} extends this with cost-model-based search over coarse-grained parallelism configurations, and \textsc{FlexFlow}~\citep{jia2019flexflow} applies simulation-guided MCMC search for training graphs. 
More recent inference-oriented frameworks, such as \textsc{AlpaServe}~\citep{yu2023alpaserve} and \textsc{Seesaw}~\citep{su2025seesaw}, optimize serving throughput or dynamically switch sharding strategies between prefill and decoding. TAPAS~\citep{shi2025tapas} identifies the large sharding search space and prunes the computation graph to accelerate the search. DFmodel~\citep{ko2024dfmodel} and UniAP~\citep{lin2025uniap} automate the parallelism strategy optimization through integer programming and COSMIC~\citep{raju2025cosmic} leverages metaheuristic algorithms like simulated annealing.
However, these automated optimization methods only target the coarse-grained strategy, neglecting the importance of the fine-grained strategy (i.e., per-operator sharding dimension). To our knowledge, our work is the first work to both explore RL-based parallelization strategy optimization with SLO constraints and co-optimize coarse and fine-grained strategies.


\section{Methodology}

\textbf{Search Framework Overview. }
We formulate the problem of optimizing the parallelization strategy as an RL search over a \textsc{MultiDiscrete} action space.
At each step, the agent proposes a joint parallelization strategy
$\boldsymbol{a} = \big(a_{\text{TP}},\, a_{\text{EP}},\, a_{\text{PP}},\, a_{\text{B}},\, \{a_{\text{dim}}^{(\ell)}\}_{\ell=1}^{L}\big),$
where $a_{\text{TP}}$, $a_{\text{EP}}$, and $a_{\text{PP}}$ denote the degrees of tensor, expert, and pipeline
parallelism, $a_{\text{B}}$ controls the batch size, and
$a_{\text{dim}}^{(\ell)} \in \{0,1,\varnothing\}$ selects the per-operator sharding dimension
for fused op $\ell$ (shard along dimension~0, dimension~1, or no sharding). Here $L$ is the number of
fused operations considered.

The environment evaluates the strategy $\boldsymbol{a}$ in our performance simulator and returns a scalar raw throughput (token/s/chip) $\text{raw}(\boldsymbol{a})$. We define the learning reward signal $r(a)$ as
\begin{equation}
  r(\boldsymbol{a}) \;=\; \alpha\,\text{raw}(\boldsymbol{a}) \;+\; \beta\,\big(\text{raw}(\boldsymbol{a}) - {b}\big),
  \label{eq:reward}
\end{equation}
where ${b}$ denotes the best raw throughput observed so far and $\alpha,\beta>0$ are scaling factors.
Thus $r(\boldsymbol{a})$ provides an improvement bonus when $\text{raw}(\boldsymbol{a})>b$ and penalizes underperforming strategies. The simulator identifies invalid configurations (e.g., SLO violations) with large negative rewards. We run for a fixed simulator call
budget $B$ and select the final strategy as the highest reward configuration encountered.

\begin{wrapfigure}{r}{0.65\textwidth}
\centering
\vspace{-15pt} 
\includegraphics[width=\linewidth]{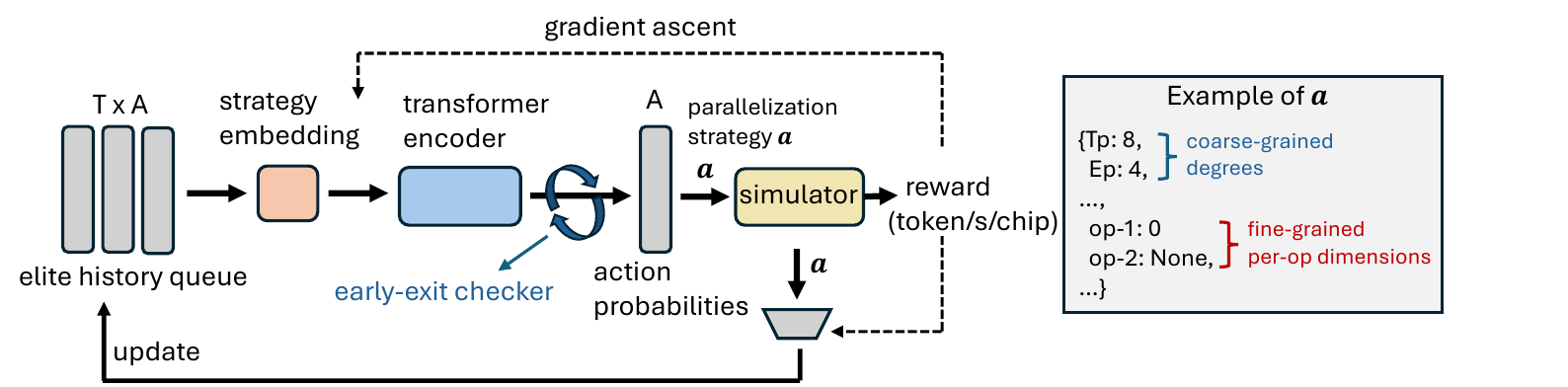}
\vspace{-15pt}
\caption{Overall search loop and the policy network design.}
\vspace{-5pt}
\label{fig:policy_network}
\end{wrapfigure}

\textbf{Policy Network. }
As shown in Fig.~\ref{fig:policy_network}, the observation $X$ to the policy is a fixed-length
history of elite strategies.
We maintain an elite history buffer (deque) of length $T$ containing the top-$T$ strategies, sorted by performance (magnitude of $r(\boldsymbol{a})$).
At each step we form $X \in \mathbb{R}^{T \times A}$ by concatenating the $T$ strategy
records (each an $A$-dimensional vector).
Each record is embedded via a linear layer ($\mathbb{R}^A \!\to\! \mathbb{R}^d$),
stacked, and passed through a Transformer encoder block, producing $Z \in \mathbb{R}^{T \times d}$.
We apply mean pooling along the history dimension,
$h=\tfrac{1}{T}\sum_{t=1}^{T} Z_{t,:}\in\mathbb{R}^{d}$,
and project $h$ with a small MLP to obtain the logits
for each sub-strategy head.
We optimize the policy with PPO to maximize expected reward of sampled strategies:
\[
L_{\pi}(\theta)=\mathbb{E}_{a\sim\pi_\theta(\cdot\mid X)}\big[r(a)\big],
\]
where the expectation is over on-policy rollouts with $a$ sampled from
$\pi_\theta(\cdot\mid X)$ given the current observation.
Every time a valid strategy is found, we update the elite history buffer if the new strategy improves
upon the worst elite in the buffer.

\textbf{Optimization and early exit. }
To save search budget and avoid local optima, we use a confidence-based early exit.
Formally, let $p_m = \pi_\theta^{(m)}(\cdot\mid X)$ denote the categorical distribution for sub-strategy head $m$.
We define the confidence of head $m$ as $\mathrm{CS}_m=\mathrm{max}(p_m)$ and
trigger an early exit if $\mathrm{CS}_m \ge \tau$ for {all} heads $m$ (e.g., $\tau=0.95$),
at which point the policy is effectively deterministic about the strategy.
To fully utilize the saved budget, we partition the total budget $B$ into smaller chunks (typically 5 chunks of equal size); upon an early exit
we {restart} the agent by reinitializing network parameters while inheriting the prior agent's $b$ and elite
history buffer.

\section{Evaluation}

\begin{wraptable}{r}{0.5\linewidth}
\centering
\vspace{-10pt} 
\begin{tabular}{lccc}
\toprule
Workloads & SA &  Ours \\
\midrule
GPT-MoE 1.2T (16k) & 1.19$\times$  & \textbf{2.76$\times$} \\
GPT-MoE 1.2T (32k) & 0.95$\times$  & \textbf{2.17$\times$}\\
GPT-MoE 1.2T (64k) & 0.71$\times$  & \textbf{2.04$\times$} \\
GPT-MoE 1.6T (16k) & 0.54$\times$  & \textbf{1.65$\times$} \\
GPT-MoE 1.6T (32k) & 0.66$\times$  & \textbf{2.31$\times$} \\
GPT-MoE 1.6T (64k) & 0.89$\times$ &\textbf{1.92$\times$} \\
\bottomrule
\end{tabular}
\caption{Normalized throughput improvement over {random walk} (RW) across models and context-lengths on H100 (mean over 10 runs; higher is better).}
\vspace{-10pt}
\label{tab:speedup_comparison}
\end{wraptable}

\textbf{Setups. } We evaluate parallelization strategy performance using an in-house roofline-based simulator\footnote{Validated against Megatron-LM heuristic parallelization strategy.}. The hardware system used in the evaluation is NVIDIA H100. We evaluate two synthetic MoE models (1.2T and 1.6T parameters)~\citep{nvidia2024gtc_slides} and consider $L=12$ fused ops; the device budget is up to 24k GPUs. This yields a search space of $\sim\!10^{9}$ joint configurations. Exhaustive evaluation would require $\sim$100 days. For all searches we allow a budget of 4000 agent forward + simulator calls (including invalid configurations), which takes less than 10 minutes. The detailed setups can be found in Appendix~\ref{sec:appendix:setup}.

\textbf{Main Results. } We compare the search quality of our proposed PPO-based agent with the other metaheuristic search algorithms, including simulated annealing (SA) and random walk (RW), which are widely used in prior works~\citep{zhou2020transferable, raju2025cosmic}, across different decoding context lengths. As shown in Table~\ref{tab:speedup_comparison}, our method consistently outperforms SA and RW under the same evaluation budget, achieving up to $2.76\times$ throughput improvement over the RW.

\begin{wrapfigure}{l}{0.5\textwidth}
\centering
\vspace{-15pt} 
\includegraphics[width=0.9\linewidth]{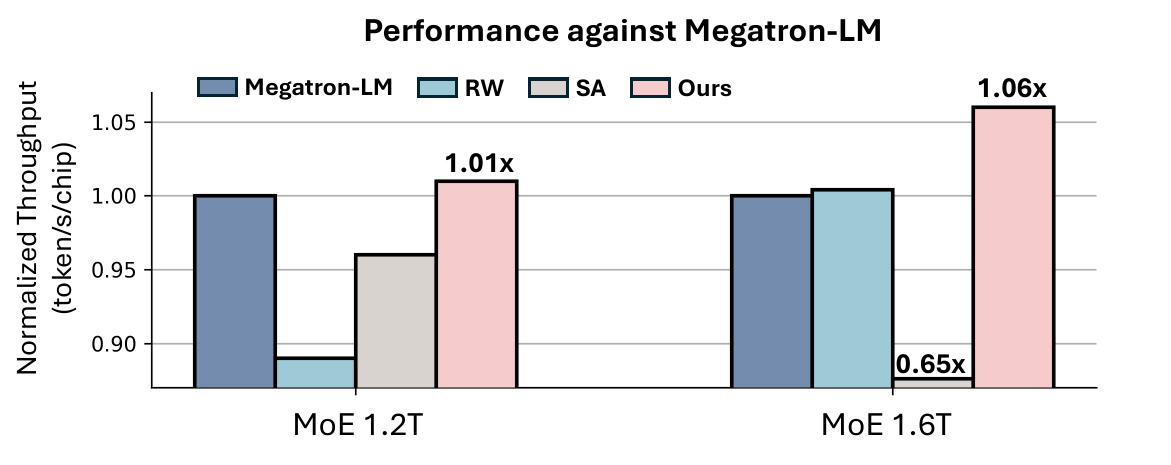}
\vspace{-5pt}
\caption{Normalized throughput of jointly optimized sharding vs. Megatron-LM heuristics (best of 10 runs; decoding, 16k context).}
\vspace{-5pt}
\label{fig:megatron}
\end{wrapfigure}

We further measure the improvement of the agent-found parallelization strategies over the heuristic ones. We set the heuristic strategy baseline to use the standard per-operator sharding dimensions from Megatron-LM~\citep{shoeybi2019megatron}. We then fix Megatron’s per-operator sharding dimensions and exhaustively search only over the parallelization degrees (TP/EP/PP/batch), which is tractable, to produce a strong heuristic baseline. The performance is measured for the decoding stage with the context length of 16k. As shown in Figure~\ref{fig:megatron}, our method consistently outperforms the Megatron-LM based heuristic parallelization strategies with up to 1.06$\times$ throughput improvement. In contrast, other metaheuristic-based agents fail to find a strategy that is better than the heuristic one. One example of the sharding strategy found by our agent is shown in Appendix~\ref{sec:appendix:example}.

\section{Conclusion and Takeaways}
\textbf{Learn to Shard} jointly optimizes parallelism degrees and per-operator sharding dimensions via a lightweight PPO-trained RL policy, achieving up to 3.5$\times$ improvement over simulated annealing and 1.06$\times$ over Megatron heuristics on H100 clusters with MoE models up to 1.6T parameters. Key takeaways include: (i) Learning-based search outperforms metaheuristics in large, sparse reward spaces under realistic constraints (HBM, interconnect, SLOs). (ii) Co-optimizing parallelization degrees and per-operator sharding enables discovery of non-standard strategies that exceed Megatron heuristics. Future work includes extending the search space to include, for example, multi-dimensional sharding.

\bibliography{refs}
\bibliographystyle{neurips_2025}
\newpage
\appendix

\section{Experimental setup}
\label{sec:appendix:setup}

We use cosine annealing for the PPO learning rate and SA temperature (initial temperature = 100). Unless noted otherwise, the elite history buffer size is $T{=}3$, the confidence threshold is $\tau{=}0.95$, and the search budget is split into $5$ equal chunks.
We use the PPO algorithm from StableBaseline3~\citep{raffin2021stable}. Step size and the rollout buffer size are both set to 2. We use 2 gradient ascents steps per update. The learning rate for PPO learning starts at $1e^{-3}$. We use a single encoder transformer block~\citep{paszke2019pytorch} in our policy network. Hidden size and strategy-embedding size are both 256.

\section{Different strategy found by agent}
\label{sec:appendix:example}

\begin{figure}[h]
\centering
\includegraphics[width=0.75\linewidth]{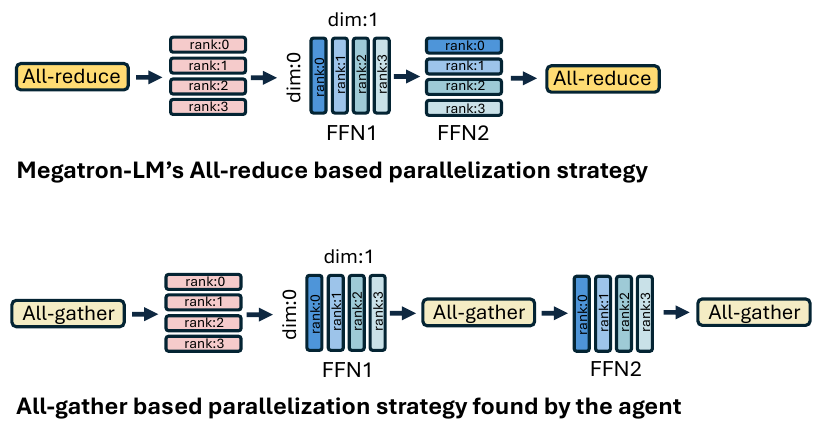}
\caption{Illustration of the all-gather based sharding strategy found by our \textbf{Learn to Shard} agent vs. Megatron-LM's all-reduce based heuristic sharding strategy. (We only show the MLP operators).}
\label{fig:appendix}
\end{figure}

Figure~\ref{fig:appendix} illustrates the difference between Megatron-LM's heuristic~\citep{shoeybi2019megatron} and one example parallelization strategy found by our agent. The main difference is that instead of sharding FFN2 operator on the feedforward dimension, the agent shards on the hidden dimension. This requires an extra all-gather communication between FFN1 and FFN2 but simplifies the original all-reduce operations into all-gather operations.

In Figure~\ref{fig:appendix}, we only show the MLP part of the fine-grained parallelization strategy. For illustrational purposes, we have a dummy coarse-grained parallelization strategy of \{TP=4, EP=1, PP=1\}.



\end{document}